# Expressing Implicit Semantic Relations without Supervision

Peter D. Turney
Institute for Information Technology
National Research Council Canada
M-50 Montreal Road
Ottawa, Ontario, Canada, K1A 0R6
peter.turney@nrc-cnrc.gc.ca

## Abstract

We present an unsupervised learning algorithm that mines large text corpora for patterns that express implicit semantic relations. For a given input word pair $X:Y$ with some unspecified semantic relations, the corresponding output list of patterns $\langle P_1,\ldots,P_m \rangle$ is ranked according to how well each pattern $P_i$ expresses the relations between $X$ and $Y$. For example, given $X$ = ostrich and $Y$ = bird, the two highest ranking output patterns are "$X$ is the largest $Y$" and "$Y$ such as the $X$". The output patterns are intended to be useful for finding further pairs with the same relations, to support the construction of lexicons, ontologies, and semantic networks. The patterns are sorted by *pertinence*, where the pertinence of a pattern $P_i$ for a word pair $X:Y$ is the expected relational similarity between the given pair and typical pairs for $P_i$. The algorithm is empirically evaluated on two tasks, solving multiple-choice SAT word analogy questions and classifying semantic relations in noun-modifier pairs. On both tasks, the algorithm achieves state-of-the-art results, performing significantly better than several alternative pattern ranking algorithms, based on tf-idf.

## 1 Introduction

In a widely cited paper, Hearst (1992) showed that the lexico-syntactic pattern "$Y$ such as the $X$" can be used to mine large text corpora for word pairs $X:Y$ in which $X$ is a hyponym (type) of $Y$. For example, if we search in a large corpus using the pattern "$Y$ such as the $X$" and we find the string "bird such as the ostrich", then we can infer that "ostrich" is a hyponym of "bird". Berland and Charniak (1999) demonstrated that the patterns "$Y$'s $X$" and "$X$ of the $Y$" can be used to mine corpora for pairs $X:Y$ in which $X$ is a meronym (part) of $Y$ (e.g., "wheel of the car").

Here we consider the *inverse* of this problem: Given a word pair $X:Y$ with some unspecified semantic relations, can we mine a large text corpus for lexico-syntactic patterns that express the implicit relations between $X$ and $Y$? For example, if we are given the pair ostrich:bird, can we discover the pattern "$Y$ such as the $X$"? We are particularly interested in discovering high quality patterns that are reliable for mining further word pairs with the same semantic relations.

In our experiments, we use a corpus of web pages containing about $5\times 10^{10}$ English words (Terra and Clarke, 2003). From co-occurrences of the pair ostrich:bird in this corpus, we can generate 516 patterns of the form "$X$ ... $Y$" and 452 patterns of the form "$Y$ ... $X$". Most of these patterns are not very useful for text mining. The main challenge is to find a way of ranking the patterns, so that patterns like "$Y$ such as the $X$" are highly ranked. Another challenge is to find a way to empirically evaluate the performance of any such pattern ranking algorithm.

For a given input word pair $X:Y$ with some unspecified semantic relations, we rank the corresponding output list of patterns $\langle P_1,\ldots,P_m \rangle$ in order of decreasing *pertinence*. The pertinence of a pattern $P_i$ for a word pair $X:Y$ is the expected relational similarity between the given pair and typical pairs that fit $P_i$. We define pertinence more precisely in Section 2.

Hearst (1992) suggests that her work may be useful for building a thesaurus. Berland and Charniak (1999) suggest their work may be useful for building a lexicon or ontology, like WordNet. Our algorithm is also applicable to these tasks. Other potential applications and related problems are discussed in Section 3.

To calculate pertinence, we must be able to measure relational similarity. Our measure is based on Latent Relational Analysis (Turney, 2005). The details are given in Section 4.

Given a word pair $X:Y$, we want our algorithm to rank the corresponding list of patterns

$\langle P_1,\ldots,P_m\rangle$ according to their value for mining text, in support of semantic network construction and similar tasks. Unfortunately, it is difficult to measure performance on such tasks. Therefore our experiments are based on two tasks that provide objective performance measures.

In Section 5, ranking algorithms are compared by their performance on solving multiple-choice SAT word analogy questions. In Section 6, they are compared by their performance on classifying semantic relations in noun-modifier pairs. The experiments demonstrate that ranking by pertinence is significantly better than several alternative pattern ranking algorithms, based on tf-idf. The performance of pertinence on these two tasks is slightly below the best performance that has been reported so far (Turney, 2005), but the difference is not statistically significant.

We discuss the results in Section 7 and conclude in Section 8.

## 2 Pertinence

The *relational similarity* between two pairs of words, $X_1:Y_1$ and $X_2:Y_2$, is the degree to which their semantic relations are analogous. For example, mason:stone and carpenter:wood have a high degree of relational similarity. Measuring relational similarity will be discussed in Section 4. For now, assume that we have a measure of the relational similarity between pairs of words, $\text{sim}_r(X_1:Y_1, X_2:Y_2) \in \Re$.

Let $W = \{X_1:Y_1,\ldots,X_n:Y_n\}$ be a set of word pairs and let $P = \{P_1,\ldots,P_m\}$ be a set of patterns. The *pertinence* of pattern $P_i$ to a word pair $X_j:Y_j$ is the *expected relational similarity* between a word pair $X_k:Y_k$, randomly selected from $W$ according to the probability distribution $p(X_k:Y_k|P_i)$, and the word pair $X_j:Y_j$:

$$\text{pertinence}(X_j:Y_j, P_i)$$
$$= \sum_{k=1}^{n} p(X_k:Y_k|P_i)\cdot \text{sim}_r(X_j:Y_j, X_k:Y_k)$$

The conditional probability $p(X_k:Y_k|P_i)$ can be interpreted as the degree to which the pair $X_k:Y_k$ is representative (i.e., *typical*) of pairs that fit the pattern $P_i$. That is, $P_i$ is pertinent to $X_j:Y_j$ if highly typical word pairs $X_k:Y_k$ for the pattern $P_i$ tend to be relationally similar to $X_j:Y_j$.

Pertinence tends to be highest with patterns that are unambiguous. The maximum value of pertinence($X_j:Y_j, P_i$) is attained when the pair $X_j:Y_j$ belongs to a cluster of highly similar pairs and the conditional probability distribution $p(X_k:Y_k|P_i)$ is concentrated on the cluster. An ambiguous pattern, with its probability spread over multiple clusters, will have less pertinence.

If a pattern with high pertinence is used for text mining, it will tend to produce word pairs that are very similar to the given word pair; this follows from the definition of pertinence. We believe this definition is the first formal measure of quality for text mining patterns.

Let $f_{k,i}$ be the number of occurrences in a corpus of the word pair $X_k:Y_k$ with the pattern $P_i$. We could estimate $p(X_k:Y_k|P_i)$ as follows:

$$p(X_k:Y_k|P_i) = f_{k,i}\bigg/\sum_{j=1}^{n} f_{j,i}$$

Instead, we first estimate $p(P_i|X_k:Y_k)$:

$$p(P_i|X_k:Y_k) = f_{k,i}\bigg/\sum_{j=1}^{m} f_{k,j}$$

Then we apply Bayes' Theorem:

$$p(X_k:Y_k|P_i) = \frac{p(X_k:Y_k)\cdot p(P_i|X_k:Y_k)}{\sum_{j=1}^{n} p(X_j:Y_j)\cdot p(P_i|X_j:Y_j)}$$

We assume $p(X_j:Y_j) = 1/n$ for all pairs in $W$:

$$p(X_k:Y_k|P_i) = p(P_i|X_k:Y_k)\bigg/\sum_{j=1}^{n} p(P_i|X_j:Y_j)$$

The use of Bayes' Theorem and the assumption that $p(X_j:Y_j) = 1/n$ for all word pairs is a way of smoothing the probability $p(X_k:Y_k|P_i)$, similar to Laplace smoothing.

## 3 Related Work

Hearst (1992) describes a method for finding patterns like *"Y such as the X"*, but her method requires human judgement. Berland and Charniak (1999) use Hearst's manual procedure.

Riloff and Jones (1999) use a mutual bootstrapping technique that can find patterns automatically, but the bootstrapping requires an initial seed of manually chosen examples for each class of words. Miller et al. (2000) propose an approach to relation extraction that was evaluated in the Seventh Message Understanding Conference (MUC7). Their algorithm requires labeled examples of each relation. Similarly, Zelenko et al. (2003) use a supervised kernel method that requires labeled training examples. Agichtein and Gravano (2000) also require training examples for each relation. Brin (1998) uses bootstrapping from seed examples of author:title pairs to discover patterns for mining further pairs.

Yangarber et al. (2000) and Yangarber (2003) present an algorithm that can find patterns automatically, but it requires an initial seed of manually designed patterns for each semantic relation. Stevenson (2004) uses WordNet to extract relations from text, but also requires initial seed patterns for each relation.

Lapata (2002) examines the task of expressing the implicit relations in nominalizations, which are noun compounds whose head noun is derived from a verb and whose modifier can be interpreted as an argument of the verb. In contrast with this work, our algorithm is not restricted to nominalizations. Section 6 shows that our algorithm works with arbitrary noun compounds and the SAT questions in Section 5 include all nine possible pairings of nouns, verbs, and adjectives.

As far as we know, our algorithm is the first unsupervised learning algorithm that can find patterns for semantic relations, given only a large corpus (e.g., in our experiments, about $5 \times 10^{10}$ words) and a moderately sized set of word pairs (e.g., 600 or more pairs in the experiments), such that the members of each pair appear together frequently in short phrases in the corpus. These word pairs are not seeds, since the algorithm does not require the pairs to be labeled or grouped; we do not assume they are homogenous.

The word pairs that we need could be generated automatically, by searching for word pairs that co-occur frequently in the corpus. However, our evaluation methods (Sections 5 and 6) both involve a predetermined list of word pairs. If our algorithm were allowed to generate its own word pairs, the overlap with the predetermined lists would likely be small. This is a limitation of our evaluation methods rather than the algorithm.

Since any two word pairs may have some relations in common and some that are not shared, our algorithm generates a unique list of patterns for each input word pair. For example, mason:stone and carpenter:wood share the pattern "$X$ carves $Y$", but the patterns "$X$ nails $Y$" and "$X$ bends $Y$" are unique to carpenter:wood. The ranked list of patterns for a word pair $X:Y$ gives the relations between $X$ and $Y$ in the corpus, sorted with the most pertinent (i.e., characteristic, distinctive, unambiguous) relations first.

Turney (2005) gives an algorithm for measuring the relational similarity between two pairs of words, called Latent Relational Analysis (LRA). This algorithm can be used to solve multiple-choice word analogy questions and to classify noun-modifier pairs (Turney, 2005), but it does not attempt to express the implicit semantic relations. Turney (2005) maps each pair $X:Y$ to a high-dimensional vector $\vec{v}$. The value of each element $v_i$ in $\vec{v}$ is based on the frequency, for the pair $X:Y$, of a corresponding pattern $P_i$. The relational similarity between two pairs, $X_1:Y_1$ and $X_2:Y_2$, is derived from the cosine of the angle between their two vectors. A limitation of this approach is that the semantic content of the vectors is difficult to interpret; the magnitude of an element $v_i$ is not a good indicator of how well the corresponding pattern $P_i$ expresses a relation of $X:Y$. This claim is supported by the experiments in Sections 5 and 6.

Pertinence (as defined in Section 2) builds on the measure of relational similarity in Turney (2005), but it has the advantage that the semantic content can be interpreted; we can point to specific patterns and say that they express the implicit relations. Furthermore, we can use the patterns to find other pairs with the same relations.

Hearst (1992) processed her text with a part-of-speech tagger and a unification-based constituent analyzer. This makes it possible to use more general patterns. For example, instead of the literal string pattern "$Y$ such as the $X$", where $X$ and $Y$ are words, Hearst (1992) used the more abstract pattern "$NP_0$ such as $NP_1$", where $NP_i$ represents a noun phrase. For the sake of simplicity, we have avoided part-of-speech tagging, which limits us to literal patterns. We plan to experiment with tagging in future work.

## 4 The Algorithm

The algorithm takes as input a set of word pairs $W = \{X_1:Y_1, \ldots, X_n:Y_n\}$ and produces as output ranked lists of patterns $\langle P_1, \ldots, P_m \rangle$ for each input pair. The following steps are similar to the algorithm of Turney (2005), with several changes to support the calculation of pertinence.

**1. Find phrases:** For each pair $X_i:Y_i$, make a list of phrases in the corpus that contain the pair. We use the Waterloo MultiText System (Clarke et al., 1998) to search in a corpus of about $5 \times 10^{10}$ English words (Terra and Clarke, 2003). Make one list of phrases that begin with $X_i$ and end with $Y_i$ and a second list for the opposite order. Each phrase must have one to three intervening words between $X_i$ and $Y_i$. The first and last words in the phrase do not need to exactly match $X_i$ and $Y_i$. The MultiText query language allows different suffixes. Veale (2004) has observed that it is easier to identify semantic relations between nouns than between other parts of speech. Therefore we use WordNet 2.0 (Miller, 1995) to guess whether $X_i$ and $Y_i$ are likely to be nouns. When they are nouns, we are relatively strict about suffixes; we only allow variation in pluralization. For all other parts of speech, we are liberal about suffixes. For example, we allow an adjective such as "inflated" to match a noun such as "inflation". With MultiText, the query "inflat*" matches both "inflated" and "inflation".

**2. Generate patterns:** For each list of phrases, generate a list of patterns, based on the phrases. Replace the first word in each phrase with the generic marker "$X$" and replace the last word with "$Y$". The intervening words in each phrase may be either left as they are or replaced with the wildcard "*". For example, the phrase "carpenter nails the wood" yields the patterns *"X nails the Y"*, *"X nails * Y"*, *"X * the Y"*, and *"X * * Y"*. Do not allow duplicate patterns in a list, but note the number of times a pattern is generated for each word pair $X_i : Y_i$ in each order ($X_i$ first and $Y_i$ last or vice versa). We call this the *pattern frequency*. It is a local frequency count, analogous to *term frequency* in information retrieval.

**3. Count pair frequency:** The *pair frequency* for a pattern is the number of lists from the preceding step that contain the given pattern. It is a global frequency count, analogous to *document frequency* in information retrieval. Note that a pair $X_i : Y_i$ yields two lists of phrases and hence two lists of patterns. A given pattern might appear in zero, one, or two of the lists for $X_i : Y_i$.

**4. Map pairs to rows:** In preparation for building a matrix **X**, create a mapping of word pairs to row numbers. For each pair $X_i : Y_i$, create a row for $X_i : Y_i$ and another row for $Y_i : X_i$. If $W$ does not already contain $\{Y_1 : X_1, \ldots, Y_n : X_n\}$, then we have effectively doubled the number of word pairs, which increases the sample size for calculating pertinence.

**5. Map patterns to columns:** Create a mapping of patterns to column numbers. For each unique pattern of the form *"X ... Y"* from Step 2, create a column for the original pattern *"X ... Y"* and another column for the same pattern with *X* and *Y* swapped, *"Y ... X"*. Step 2 can generate millions of distinct patterns. The experiment in Section 5 results in 1,706,845 distinct patterns, yielding 3,413,690 columns. This is too many columns for matrix operations with today's standard desktop computer. Most of the patterns have a very low pair frequency. For the experiment in Section 5, 1,371,702 of the patterns have a pair frequency of one. To keep the matrix **X** manageable, we drop all patterns with a pair frequency less than ten. For Section 5, this leaves 42,032 patterns, yielding 84,064 columns. Turney (2005) limited the matrix to 8,000 columns, but a larger pool of patterns is better for our purposes, since it increases the likelihood of finding good patterns for expressing the semantic relations of a given word pair.

**6. Build a sparse matrix:** Build a matrix **X** in sparse matrix format. The value for the cell in row *i* and column *j* is the pattern frequency of the *j*-th pattern for the the *i*-th word pair.

**7. Calculate entropy:** Apply log and entropy transformations to the sparse matrix **X** (Landauer and Dumais, 1997). Each cell is replaced with its logarithm, multiplied by a weight based on the negative entropy of the corresponding column vector in the matrix. This gives more weight to patterns that vary substantially in frequency for each pair.

**8. Apply SVD:** After log and entropy transforms, apply the Singular Value Decomposition (SVD) to **X** (Golub and Van Loan, 1996). SVD decomposes **X** into a product of three matrices $\mathbf{U}\Sigma\mathbf{V}^T$, where **U** and **V** are in column orthonormal form (i.e., the columns are orthogonal and have unit length) and $\Sigma$ is a diagonal matrix of singular values (hence SVD). If **X** is of rank $r$, then $\Sigma$ is also of rank $r$. Let $\Sigma_k$, where $k < r$, be the diagonal matrix formed from the top $k$ singular values, and let $\mathbf{U}_k$ and $\mathbf{V}_k$ be the matrices produced by selecting the corresponding columns from **U** and **V**. The matrix $\mathbf{U}_k\Sigma_k\mathbf{V}_k^T$ is the matrix of rank $k$ that best approximates the original matrix **X**, in the sense that it minimizes the approximation errors (Golub and Van Loan, 1996). Following Landauer and Dumais (1997), we use $k = 300$. We may think of this matrix $\mathbf{U}_k\Sigma_k\mathbf{V}_k^T$ as a smoothed version of the original matrix. SVD is used to reduce noise and compensate for sparseness (Landauer and Dumais, 1997).

**9. Calculate cosines:** The relational similarity between two pairs, $\text{sim}_r(X_1 : Y_1, X_2 : Y_2)$, is given by the cosine of the angle between their corresponding row vectors in the matrix $\mathbf{U}_k\Sigma_k\mathbf{V}_k^T$ (Turney, 2005). To calculate pertinence, we will need the relational similarity between all possible pairs of pairs. All of the cosines can be efficiently derived from the matrix $\mathbf{U}_k\Sigma_k(\mathbf{U}_k\Sigma_k)^T$ (Landauer and Dumais, 1997).

**10. Calculate conditional probabilities:** Using Bayes' Theorem (see Section 2) and the raw frequency data in the matrix **X** from Step 6, before log and entropy transforms, calculate the conditional probability $p(X_i : Y_i | P_j)$ for every row (word pair) and every column (pattern).

**11. Calculate pertinence:** With the cosines from Step 9 and the conditional probabilities from Step 10, calculate pertinence($X_i : Y_i, P_j$) for every row $X_i : Y_i$ and every column $P_j$ for which $p(X_i : Y_i | P_j) > 0$. When $p(X_i : Y_i | P_j) = 0$, it is possible that pertinence($X_i : Y_i, P_j$) > 0, but we avoid calculating pertinence in these cases for two reasons. First, it speeds computation, because **X** is sparse, so $p(X_i : Y_i | P_j) = 0$ for most rows and columns. Second, $p(X_i : Y_i | P_j) = 0$ implies that the pattern $P_j$ does not actually appear with the word pair $X_i : Y_i$ in the corpus; we are only guessing that the pattern is appropriate for the word pair, and we could be wrong. Therefore we prefer to limit ourselves to patterns and word pairs that have actually been observed in the corpus. For each pair $X_i : Y_i$ in *W*, output two separate ranked lists, one for patterns of the form *"X … Y"* and another for patterns of the form

"Y ... X", where the patterns in both lists are sorted in order of decreasing pertinence to $X_i:Y_i$. Ranking serves as a kind of normalization. We have found that the relative rank of a pattern is more reliable as an indicator of its importance than the absolute pertinence. This is analogous to information retrieval, where documents are ranked in order of their relevance to a query. The relative rank of a document is more important than its actual numerical score (which is usually hidden from the user of a search engine). Having two separate ranked lists helps to avoid bias. For example, ostrich:bird generates 516 patterns of the form "X ... Y" and 452 patterns of the form "Y ... X". Since there are more patterns of the form "X ... Y", there is a slight bias towards these patterns. If the two lists were merged, the "Y ... X" patterns would be at a disadvantage.

## 5 Experiments with Word Analogies

In these experiments, we evaluate pertinence using 374 college-level multiple-choice word analogies, taken from the SAT test. For each question, there is a target word pair, called the *stem* pair, and five *choice* pairs. The task is to find the choice that is most analogous (i.e., has the highest relational similarity) to the stem. This choice pair is called the *solution* and the other choices are *distractors*. Since there are six word pairs per question (the stem and the five choices), there are $374 \times 6 = 2244$ pairs in the input set $W$. In Step 4 of the algorithm, we double the pairs, but we also drop some pairs because they do not co-occur in the corpus. This leaves us with 4194 rows in the matrix. As mentioned in Step 5, the matrix has 84,064 columns (patterns). The sparse matrix density is 0.91%.

To answer a SAT question, we generate ranked lists of patterns for each of the six word pairs. Each choice is evaluated by taking the intersection of its patterns with the stem's patterns. The shared patterns are scored by the average of their rank in the stem's lists and the choice's lists. Since the lists are sorted in order of decreasing pertinence, a low score means a high pertinence. Our guess is the choice with the lowest scoring shared pattern.

Table 1 shows three examples, two questions that are answered correctly followed by one that is answered incorrectly. The correct answers are in bold font. For the first question, the stem is ostrich:bird and the best choice is (a) lion:cat. The highest ranking pattern that is shared by both of these pairs is "Y such as the X". The third question illustrates that, even when the answer is incorrect, the best shared pattern ("Y powered * * X") may be plausible.

| Word pair | Best shared pattern | Score |
|---|---|---|
| 1. ostrich:bird | | |
| **(a) lion:cat** | **"Y such as the X"** | **1.0** |
| (b) goose:flock | "X * * breeding Y" | 43.5 |
| (c) ewe:sheep | "X are the only Y" | 13.5 |
| (d) cub:bear | "Y are called X" | 29.0 |
| (e) primate:monkey | "Y is the * X" | 80.0 |
| 2. traffic:street | | |
| (a) ship:gangplank | "X * down the Y" | 53.0 |
| (b) crop:harvest | "X * adjacent * Y" | 248.0 |
| (c) car:garage | "X * a residential Y" | 63.0 |
| (d) pedestrians:feet | "Y * accommodate X" | 23.0 |
| **(e) water:riverbed** | **"Y that carry X"** | **17.0** |
| 3. locomotive:train | | |
| (a) horse:saddle | "X carrying * Y" | 82.0 |
| **(b) tractor:plow** | **"X pulled * Y"** | **7.0** |
| (c) rudder:rowboat | "Y * X" | 319.0 |
| (d) camel:desert | "Y with two X" | 43.0 |
| (e) gasoline:automobile | "Y powered * * X" | 5.0 |

Table 1. Three examples of SAT questions.

Table 2 shows the four highest ranking patterns for the stem and solution for the first example. The pattern "X lion Y" is anomalous, but the other patterns seem reasonable. The shared pattern "Y such as the X" is ranked 1 for both pairs, hence the average score for this pattern is 1.0, as shown in Table 1. Note that the "ostrich is the largest bird" and "lions are large cats", but the largest cat is the Siberian tiger.

| Word pair | "X ... Y" | "Y ... X" |
|---|---|---|
| ostrich:bird | "X is the largest Y" | **"Y such as the X"** |
| | "X is * largest Y" | "Y such * the X" |
| lion:cat | "X lion Y" | **"Y such as the X"** |
| | "X are large Y" | "Y and mountain X" |

Table 2. The highest ranking patterns.

Table 3 lists the top five pairs in $W$ that match the pattern "Y such as the X". The pairs are sorted by $p(X:Y|P)$. The pattern "Y such as the X" is one of 146 patterns that are shared by ostrich:bird and lion:cat. Most of these shared patterns are not very informative.

| Word pair | Conditional probability |
|---|---|
| heart:organ | 0.49342 |
| dodo:bird | 0.08888 |
| elbow:joint | 0.06385 |
| ostrich:bird | 0.05774 |
| semaphore:signal | 0.03741 |

Table 3. The top five pairs for "Y such as the X".

In Table 4, we compare ranking patterns by pertinence to ranking by various other measures, mostly based on varieties of tf-idf (term frequency times inverse document frequency, a common way to rank documents in information retrieval). The tf-idf measures are taken from Salton and Buckley (1988). For comparison, we also include three algorithms that do not rank patterns (the bottom three rows in the table). These three algorithms can answer the SAT questions, but they do not provide any kind of explanation for their answers.

|   | Algorithm | Prec. | Rec. | F |
|---|---|---|---|---|
| 1 | pertinence (Step 11) | 55.7 | 53.5 | 54.6 |
| 2 | log and entropy matrix (Step 7) | 43.5 | 41.7 | 42.6 |
| 3 | TF = $f$, IDF = $\log((N-n)/n)$ | 43.2 | 41.4 | 42.3 |
| 4 | TF = $\log(f+1)$, IDF = $\log(N/n)$ | 42.9 | 41.2 | 42.0 |
| 5 | TF = $f$, IDF = $\log(N/n)$ | 42.9 | 41.2 | 42.0 |
| 6 | TF = $\log(f+1)$, IDF = $\log((N-n)/n)$ | 42.3 | 40.6 | 41.4 |
| 7 | TF = 1.0, IDF = $1/n$ | 41.5 | 39.8 | 40.6 |
| 8 | TF = $f$, IDF = $1/n$ | 41.5 | 39.8 | 40.6 |
| 9 | TF = $0.5 + 0.5 * (f/F)$, IDF = $\log(N/n)$ | 41.5 | 39.8 | 40.6 |
| 10 | TF = $\log(f+1)$, IDF = $1/n$ | 41.2 | 39.6 | 40.4 |
| 11 | $p(X{:}Y|P)$ (Step 10) | 39.8 | 38.2 | 39.0 |
| 12 | SVD matrix (Step 8) | 35.9 | 34.5 | 35.2 |
| 13 | random | 27.0 | 25.9 | 26.4 |
| 14 | TF = $1/f$, IDF = 1.0 | 26.7 | 25.7 | 26.2 |
| 15 | TF = $f$, IDF = 1.0 (Step 6) | 18.1 | 17.4 | 17.7 |
| 16 | Turney (2005) | 56.8 | 56.1 | 56.4 |
| 17 | Turney and Littman (2005) | 47.7 | 47.1 | 47.4 |
| 18 | Veale (2004) | 42.8 | 42.8 | 42.8 |

Table 4. Performance of various algorithms on SAT.

All of the pattern ranking algorithms are given exactly the same sets of patterns to rank. Any differences in performance are due to the ranking method alone. The algorithms may skip questions when the word pairs do not co-occur in the corpus. All of the ranking algorithms skip the same set of 15 of the 374 SAT questions. *Precision* is defined as the percentage of correct answers out of the questions that were answered (not skipped). *Recall* is the percentage of correct answers out of the maximum possible number correct (374). The F measure is the harmonic mean of precision and recall.

For the tf-idf methods in Table 4, $f$ is the pattern frequency, $n$ is the pair frequency, $F$ is the maximum $f$ for all patterns for the given word pair, and $N$ is the total number of word pairs. By "TF = $f$, IDF = $1/n$", for example (row 8), we mean that $f$ plays a role that is analogous to term frequency and $1/n$ plays a role that is analogous to inverse document frequency. That is, in row 8, the patterns are ranked in decreasing order of pattern frequency divided by pair frequency.

Table 4 also shows some ranking methods based on intermediate calculations in the algorithm in Section 4. For example, row 2 in Table 4 gives the results when patterns are ranked in order of decreasing values in the corresponding cells of the matrix **X** from Step 7.

Row 12 in Table 4 shows the results we would get using Latent Relational Analysis (Turney, 2005) to rank patterns. The results in row 12 support the claim made in Section 3, that LRA is not suitable for ranking patterns, although it works well for answering the SAT questions (as we see in row 16). The vectors in LRA yield a good measure of relational similarity, but the magnitude of the value of a specific element in a vector is not a good indicator of the quality of the corresponding pattern.

The best method for ranking patterns is pertinence (row 1 in Table 4). As a point of comparison, the performance of the average senior highschool student on the SAT analogies is about 57% (Turney and Littman, 2005). The second best method is to use the values in the matrix **X** after the log and entropy transformations in Step 7 (row 2). The difference between these two methods is statistically significant with 95% confidence. Pertinence (row 1) performs slightly below Latent Relational Analysis (row 16; Turney, 2005), but the difference is not significant.

Randomly guessing answers should yield an F of 20% (1 out of 5 choices), but ranking patterns randomly (row 13) results in an F of 26.4%. This is because the stem pair tends to share more patterns with the solution pair than with the distractors. The minimum of a large set of random numbers is likely to be lower than the minimum of a small set of random numbers.

## 6 Experiments with Noun-Modifiers

In these experiments, we evaluate pertinence on the task of classifying noun-modifier pairs. The problem is to classify a noun-modifier pair, such as "flu virus", according to the semantic relation between the head noun (virus) and the modifier (flu). For example, "flu virus" is classified as a *causality* relation (the flu is *caused by* a virus). For these experiments, we use a set of 600 manually labeled noun-modifier pairs (Nastase and Szpakowicz, 2003). There are five general classes of labels with thirty subclasses. We present here the results with five classes; the results with thirty subclasses follow the same trends (that is, pertinence performs significantly better than the other ranking methods). The five classes are *causality* (storm cloud), *temporality* (daily exercise), *spatial* (desert storm), *participant* (student protest), and *quality* (expensive book).

The input set $W$ consists of the 600 noun-modifier pairs. This set is doubled in Step 4, but we drop some pairs because they do not co-occur in the corpus, leaving us with 1184 rows in the matrix. There are 16,849 distinct patterns with a pair frequency of ten or more, resulting in 33,698 columns. The matrix density is 2.57%.

To classify a noun-modifier pair, we use a single nearest neighbour algorithm with leave-one-out cross-validation. We split the set 600 times. Each pair gets a turn as the single testing example, while the other 599 pairs serve as training examples. The testing example is classified according to the label of its nearest neighbour in the training set. The distance between two noun-modifier pairs is measured by the average rank of their best shared pattern. Table 5 shows the resulting precision, recall, and F, when ranking patterns by pertinence.

| Class name | Prec. | Rec. | F | Class size |
|---|---|---|---|---|
| causality | 37.3 | 36.0 | 36.7 | 86 |
| participant | 61.1 | 64.4 | 62.7 | 260 |
| quality | 49.3 | 50.7 | 50.0 | 146 |
| spatial | 43.9 | 32.7 | 37.5 | 56 |
| temporality | 64.7 | 63.5 | 64.1 | 52 |
| all | 51.3 | 49.5 | 50.2 | 600 |

Table 5. Performance on noun-modifiers.

To gain some insight into the algorithm, we examined the 600 best shared patterns for each pair and its single nearest neighbour. For each of the five classes, Table 6 lists the most frequent pattern among the best shared patterns for the given class. All of these patterns seem appropriate for their respective classes.

| Class | Most frequent pattern | Example pair |
|---|---|---|
| causality | "Y * causes X" | "cold virus" |
| participant | "Y of his X" | "dream analysis" |
| quality | "Y made of X" | "copper coin" |
| spatial | "X * * terrestrial Y" | "aquatic mammal" |
| temporality | "Y in * early X" | "morning frost" |

Table 6. Most frequent of the best shared patterns.

Table 7 gives the performance of pertinence on the noun-modifier problem, compared to various other pattern ranking methods. The bottom two rows are included for comparison; they are not pattern ranking algorithms. The best method for ranking patterns is pertinence (row 1 in Table 7). The difference between pertinence and the second best ranking method (row 2) is statistically significant with 95% confidence. Latent Relational Analysis (row 16) performs slightly better than pertinence (row 1), but the difference is not statistically significant.

Row 6 in Table 7 shows the results we would get using Latent Relational Analysis (Turney, 2005) to rank patterns. Again, the results support the claim in Section 3, that LRA is not suitable for ranking patterns. LRA can classify the noun-modifiers (as we see in row 16), but it cannot express the implicit semantic relations that make an unlabeled noun-modifier in the testing set similar to its nearest neighbour in the training set.

| | Algorithm | Prec. | Rec. | F |
|---|---|---|---|---|
| 1 | pertinence (Step 11) | 51.3 | 49.5 | 50.2 |
| 2 | TF = log(f+1), IDF = 1/n | 37.4 | 36.5 | 36.9 |
| 3 | TF = log(f+1), IDF = log(N/n) | 36.5 | 36.0 | 36.2 |
| 4 | TF = log(f+1), IDF = log((N-n)/n) | 36.0 | 35.4 | 35.7 |
| 5 | TF = f, IDF = log((N-n)/n) | 36.0 | 35.3 | 35.6 |
| 6 | SVD matrix (Step 8) | 43.9 | 33.4 | 34.8 |
| 7 | TF = f, IDF = 1/n | 35.4 | 33.6 | 34.3 |
| 8 | log and entropy matrix (Step 7) | 35.6 | 33.3 | 34.1 |
| 9 | TF = f, IDF = log(N/n) | 34.1 | 31.4 | 32.2 |
| 10 | TF = 0.5 + 0.5 * (f/F), IDF = log(N/n) | 31.9 | 31.7 | 31.6 |
| 11 | p(X:Y|P) (Step 10) | 31.8 | 30.8 | 31.2 |
| 12 | TF = 1.0, IDF = 1/n | 29.2 | 28.8 | 28.7 |
| 13 | random | 19.4 | 19.3 | 19.2 |
| 14 | TF = 1/f, IDF = 1.0 | 20.3 | 20.7 | 19.2 |
| 15 | TF = f, IDF = 1.0 (Step 6) | 12.8 | 19.7 | 8.0 |
| 16 | Turney (2005) | 55.9 | 53.6 | 54.6 |
| 17 | Turney and Littman (2005) | 43.4 | 43.1 | 43.2 |

Table 7. Performance on noun-modifiers.

## 7 Discussion

Computing pertinence took about 18 hours for the experiments in Section 5 and 9 hours for Section 6. In both cases, the majority of the time was spent in Step 1, using MultiText (Clarke et al., 1998) to search through the corpus of $5 \times 10^{10}$ words. MultiText was running on a Beowulf cluster with sixteen 2.4 GHz Intel Xeon CPUs. The corpus and the search index require about one terabyte of disk space. This may seem computationally demanding by today's standards, but progress in hardware will soon allow an average desktop computer to handle corpora of this size.

Although the performance on the SAT analogy questions (54.6%) is near the level of the average senior highschool student (57%), there is room for improvement. For applications such as building a thesaurus, lexicon, or ontology, this level of performance suggests that our algorithm could assist, but not replace, a human expert.

One possible improvement would be to add part-of-speech tagging or parsing. We have done some preliminary experiments with parsing and plan to explore tagging as well. A difficulty is that much of the text in our corpus does not consist of properly formed sentences, since the text comes from web pages. This poses problems for most part-of-speech taggers and parsers.

## 8 Conclusion

Latent Relational Analysis (Turney, 2005) provides a way to measure the relational similarity between two word pairs, but it gives us little insight into how the two pairs are similar. In effect,

LRA is a black box. The main contribution of this paper is the idea of pertinence, which allows us to take an opaque measure of relational similarity and use it to find patterns that express the implicit semantic relations between two words.

The experiments in Sections 5 and 6 show that ranking patterns by pertinence is superior to ranking them by a variety of tf-idf methods. On the word analogy and noun-modifier tasks, pertinence performs as well as the state-of-the-art, LRA, but pertinence goes beyond LRA by making relations explicit.

## Acknowledgements

Thanks to Joel Martin, David Nadeau, and Deniz Yuret for helpful comments and suggestions.

## References

Eugene Agichtein and Luis Gravano. 2000. Snowball: Extracting relations from large plain-text collections. In *Proceedings of the Fifth ACM Conference on Digital Libraries (ACM DL 2000)*, pages 85-94.

Matthew Berland and Eugene Charniak. 1999. Finding parts in very large corpora. In *Proceedings of the 37th Annual Meeting of the Association for Computational Linguistics (ACL-99)*, pages 57-64.

Sergey Brin. 1998. Extracting patterns and relations from the World Wide Web. In *WebDB Workshop at the 6th International Conference on Extending Database Technology (EDBT-98)*, pages 172-183.

Charles L.A. Clarke, Gordon V. Cormack, and Christopher R. Palmer. 1998. An overview of MultiText. *ACM SIGIR Forum*, 32(2):14-15.

Gene H. Golub and Charles F. Van Loan. 1996. *Matrix Computations*. Third edition. Johns Hopkins University Press, Baltimore, MD.

Marti A. Hearst. 1992. Automatic acquisition of hyponyms from large text corpora. In *Proceedings of the 14th International Conference on Computational Linguistics (COLING-92)*, pages 539-545.

Thomas K. Landauer and Susan T. Dumais. 1997. A solution to Plato's problem: The latent semantic analysis theory of the acquisition, induction, and representation of knowledge. *Psychological Review*, 104(2):211-240.

Maria Lapata. 2002. The disambiguation of nominalisations. *Computational Linguistics*, 28(3):357-388.

George A. Miller. 1995. WordNet: A lexical database for English. *Communications of the ACM*, 38(11):39-41.

Scott Miller, Heidi Fox, Lance Ramshaw, and Ralph Weischedel. 2000. A novel use of statistical parsing to extract information from text. In *Proceedings of the Sixth Applied Natural Language Processing Conference (ANLP 2000)*, pages 226-233.

Vivi Nastase and Stan Szpakowicz. 2003. Exploring noun-modifier semantic relations. In *Fifth International Workshop on Computational Semantics (IWCS-5)*, pages 285-301.

Ellen Riloff and Rosie Jones. 1999. Learning dictionaries for information extraction by multi-level bootstrapping. In *Proceedings of the 16th National Conference on Artificial Intelligence (AAAI-99)*, pages 474-479.

Gerard Salton and Chris Buckley. 1988. Term weighting approaches in automatic text retrieval. *Information Processing and Management*, 24(5):513-523.

Mark Stevenson. 2004. An unsupervised WordNet-based algorithm for relation extraction. In *Proceedings of the Fourth International Conference on Language Resources and Evaluation (LREC) Workshop, Beyond Named Entity Recognition: Semantic Labelling for NLP Tasks*, Lisbon, Portugal.

Egidio Terra and Charles L.A. Clarke. 2003. Frequency estimates for statistical word similarity measures. In *Proceedings of the Human Language Technology and North American Chapter of Association of Computational Linguistics Conference (HLT/NAACL-03)*, pages 244-251.

Peter D. Turney. 2005. Measuring semantic similarity by latent relational analysis. In *Proceedings of the Nineteenth International Joint Conference on Artificial Intelligence (IJCAI-05)*, pages 1136-1141.

Peter D. Turney and Michael L. Littman. 2005. Corpus-based learning of analogies and semantic relations. *Machine Learning*, 60(1-3):251-278.

Tony Veale. 2004. WordNet sits the SAT: A knowledge-based approach to lexical analogy. In *Proceedings of the 16th European Conference on Artificial Intelligence (ECAI 2004)*, pages 606-612.

Roman Yangarber, Ralph Grishman, Pasi Tapanainen, and Silja Huttunen. 2000. Unsupervised discovery of scenario-level patterns for information extraction. In *Proceedings of the Sixth Applied Natural Language Processing Conference (ANLP 2000)*, pages 282-289.

Roman Yangarber. 2003. Counter-training in discovery of semantic patterns. In *Proceedings of the 41st Annual Meeting of the Association for Computational Linguistics (ACL-03)*, pages 343-350.

Dmitry Zelenko, Chinatsu Aone, and Anthony Richardella. 2003. Kernel methods for relation extraction. *Journal of Machine Learning Research*, 3:1083-1106.